\documentclass[10pt,twocolumn,letterpaper]{article}

\usepackage[pagenumbers]{cvpr} 

\usepackage[dvipsnames]{xcolor}

\usepackage{xcolor,colortbl}
\usepackage[dvipsnames]{xcolor}
\usepackage{adjustbox}
\usepackage{ulem}
\usepackage[symbol]{footmisc}
\usepackage{cuted}
\usepackage{capt-of}
\usepackage{graphicx}
\usepackage[accsupp]{axessibility}

\definecolor{tabfirst}{rgb}{1, 0.7, 0.7} 
\definecolor{tabsecond}{rgb}{1, 0.85, 0.7} 
\definecolor{tabthird}{rgb}{1, 1, 0.7} 

\usepackage{xcolor,pifont}
\newcommand*\colourcheck[1]{%
  \expandafter\newcommand\csname #1check\endcsname{\textcolor{#1}{\ding{52}}}%
}
\newcommand*\colourcross[1]{%
  \expandafter\newcommand\csname #1cross\endcsname{\textcolor{#1}{\ding{55}}}%
}
\colourcheck{green}
\colourcross{red}

\definecolor{cvprblue}{rgb}{0.21,0.49,0.74}
\usepackage[pagebackref,breaklinks,colorlinks,citecolor=cvprblue]{hyperref}

\def\paperID{23} 
\def\confName{CVPR}
\def\confYear{2024}

\title{EdgeRelight360: Text-Conditioned 360-Degree HDR Image Generation for Real-Time On-Device Video Portrait Relighting}

\author{ Min-Hui Lin\thanks{denotes equal contribution.} $\hspace{0.5cm}$ Mahesh Reddy\footnotemark[1] $\hspace{0.5cm}$  Guillaume Berger $\hspace{0.5cm}$ Michel Sarkis $\hspace{0.5cm}$ Fatih Porikli $\hspace{0.5cm}$ Ning Bi\\
Qualcomm AI Research\thanks{Qualcomm AI Research is an initiative of Qualcomm Technologies, Inc.}\\
}

\begin{document}
\maketitle
\begin{abstract}
In this paper, we present EdgeRelight360, an approach for real-time video portrait relighting on mobile devices, utilizing text-conditioned generation of 360-degree high dynamic range image (HDRI) maps. Our method proposes a diffusion-based text-to-360-degree image generation in the HDR domain, taking advantage of the HDR10 standard. This technique facilitates the generation of high-quality, realistic lighting conditions from textual descriptions, offering flexibility and control in portrait video relighting task. Unlike the previous relighting frameworks, our proposed system performs video relighting directly on-device, enabling real-time inference with real 360-degree HDRI maps. This on-device processing ensures both privacy and guarantees low runtime, providing an immediate response to changes in lighting conditions or user inputs. Our approach paves the way for new possibilities in real-time video applications, including video conferencing, gaming, and augmented reality, by allowing dynamic, text-based control of lighting conditions.
\end{abstract}    
\section{Introduction}

\label{sec:intro}

On-device video conferencing has emerged as a vital tool in our daily communications. 
Customizing the background in these platforms enhances user privacy, yet the default backgrounds are usually confined to pre-existing 2D images, and the mismatch in lighting between the subject and the virtual background often compromises the quality of the immersive experience.

In recent years, generative content creation, image/video relighting, and edge computing have witnessed a surge in quality and interest. In this context, we propose an on-device inference pipeline involving the generation of a 360-degree high dynamic range image (HDRI) map from a textual description provided by the user, followed by portrait relighting to seamlessly integrate a streaming user into the newly generated scene. 

\begin{figure}
    \centering
    \includegraphics[width=0.47\textwidth]{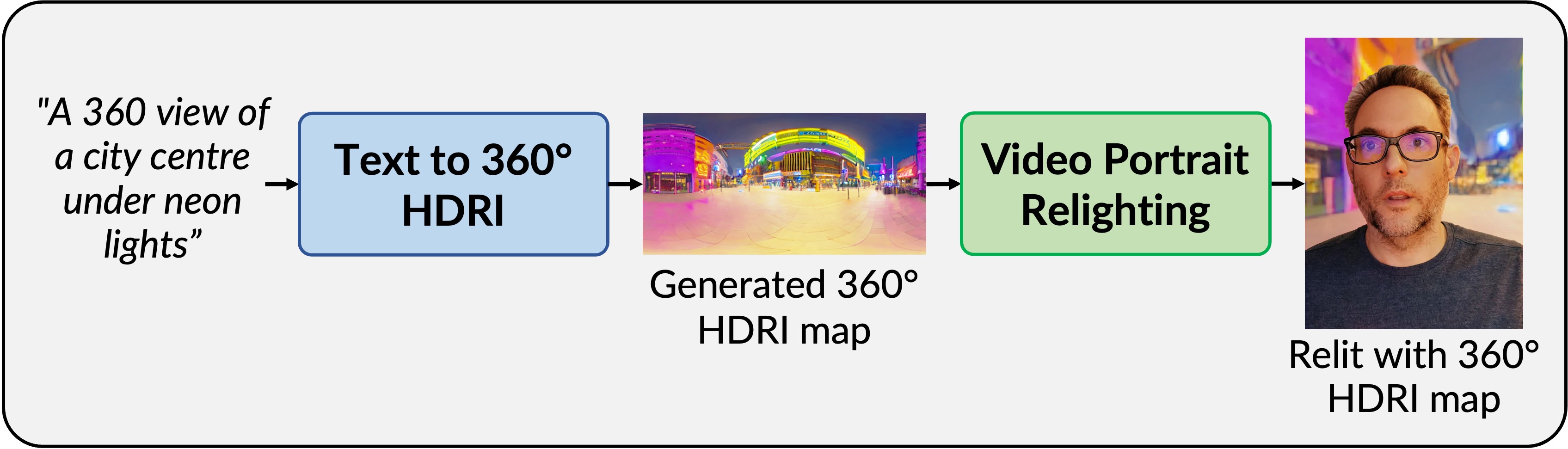}
    \caption{Our proposed method for generating 360-degree environment map from text prompt followed by video portrait relighting in real-time on mobile devices.}
    \label{fig:method_overview}
\end{figure}


To enable efficient and high-quality on-device deployments, there are several challenges in 360-degree HDRI map generation and video relighting, respectively. HDRI maps play a pivotal role in creating backgrounds and lighting for immersive on-device applications. Although an existing approach~\cite{chen2022text2light} can generate high-resolution HDRI maps, 
this method involves a complex two-stage setup which first generates a low dynamic range (LDR) 360-degree image from text, before transforming it into an HDRI map in the second stage. Unfortunately, the complexity of the neural network designs used in this approach hinder its implementation on compute-constrained devices, and the adversarial training framework often results in a lack of diversity in the generated HDRI environment maps~\cite{chen2022text2light}.

Relighting helps to naturally embed captured subjects into new environments by synthesizing physically consistent lighting effects. To cope with dynamic input lighting and to better synthesize high-frequency lighting effects, recent relighting methods \cite{nestmeyer2020learning, wang2020single, pandey2021total, hou2022face, song2022real, qiu2024relitalk} first perform learning-based intrinsic decomposition to obtain surface normal and albedo, and then incorporate per-pixel lighting representations as explicit priors into the network design. Despite producing promising relighting results, their complex structures, involving multiple networks or decoders, hinder deployment on compute-constrained devices. Additionally, current state-of-the-art relighting approaches~\cite{pandey2021total} exhibit issues with temporal stability, particularly in clothing regions. 


In our work, we address the above challenges and propose a portrait video relighting framework at the edge using text-conditioned generations of 360-degree HDRI maps. As illustrated in Fig.~\ref{fig:method_overview}, we generate a 360-degree HDRI map with real-world brightness information from a text prompt. 
To achieve this, we leverage the generative capabilities of a Stable Diffusion~\cite{rombach2021high} model to produce 360-degree HDRI maps by training it on 8-bit quantized HDRI maps following the HDR10 standard~\cite{standard2014high, series2012parameter}. 

Following this, we use the generated 360-degree image as an omnidirectional illumination source in the proposed portrait video relighting pipeline. A key novelty of our relighting model is the balance between relighting performance and computational efficiency. 
To enhance the computational efficiency, we only leverage a single network to infer surface normals for computing the diffuse light map and propose a shading equation for relighting.
And to ensure better relighting performance, we propose to add diffuse shading to the input camera images for realism, and apply a temporal filter to enhance temporal consistency.

Finally, we compute a diffuse light map from the generated HDRI map for the current view specified by the user, apply lighting effects to the subject and composite the relit portrait with the new background created from the panorama.

In summary, our main contributions include:
\begin{itemize}
    \item We propose an end-to-end on-device framework to generate 360-degree HDRI maps from text descriptions, and use it to relight video portraits in real-time. By combining HDRI map generation and lightweight video relighting, users can virtually appear anywhere imaginable in video applications at the edge.
    \item We present a diffusion-based text-conditioned 360-degree HDRI map generation to produce diverse environment maps 
    on-device for relighting video portraits.
    \item We propose a light-weight video relighting framework combining a normal estimation network and light adding based rendering. Our on-device implementation shows realistic, fast, and stable relighting results for in-the-wild portrait videos, demonstrating the effectiveness, efficiency, temporal consistency, and generalization of the proposed method.
\end{itemize}
\section{Related work}

\begin{figure}[!t]
    \centering
    \includegraphics[width=0.4\textwidth]{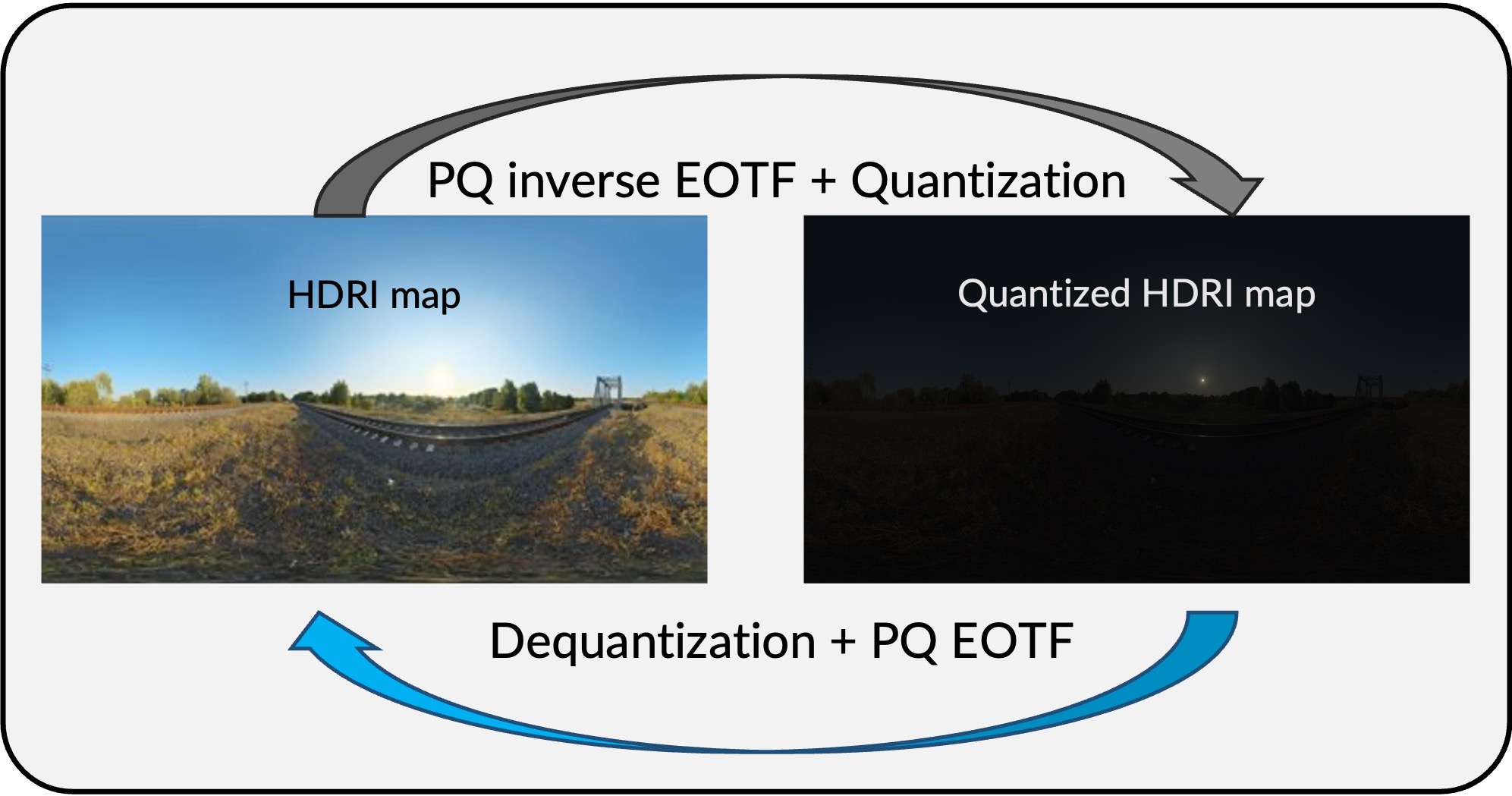}
    \caption{We propose to combine PQ inverse EOTF used in the HDR10 standard with 8-bit quantization to obtain the quantized HDRI maps to generate the training dataset. Similarly, dequantization and the PQ EOTF can be performed to recover the original HDRI map.}
    \label{fig:pq_eotf}
\end{figure}

\begin{figure*}[!t]
    \centering
    \includegraphics[width=0.75\textwidth]{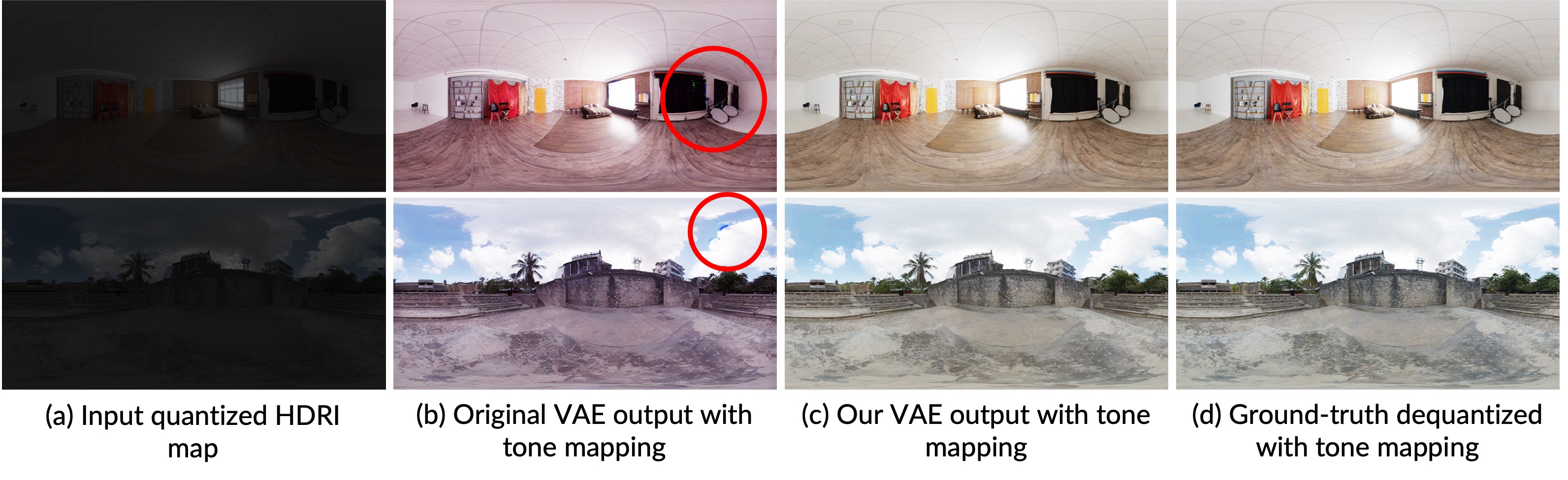}
    \caption{Encoding and decoding (a) quantized HDRI map with a pre-trained VAE leads to significant artifacts such as (b) blue patches and RGB color distortions. However, these issues can be resolved by (c) fine-tuning the VAE on quantized HDRI maps, which reproduces the quantized images close to the (d) dequantized original images.}
    \label{fig:vae}
\end{figure*}

\begin{figure}[!t]
    \centering
    \includegraphics[width=0.4\textwidth]{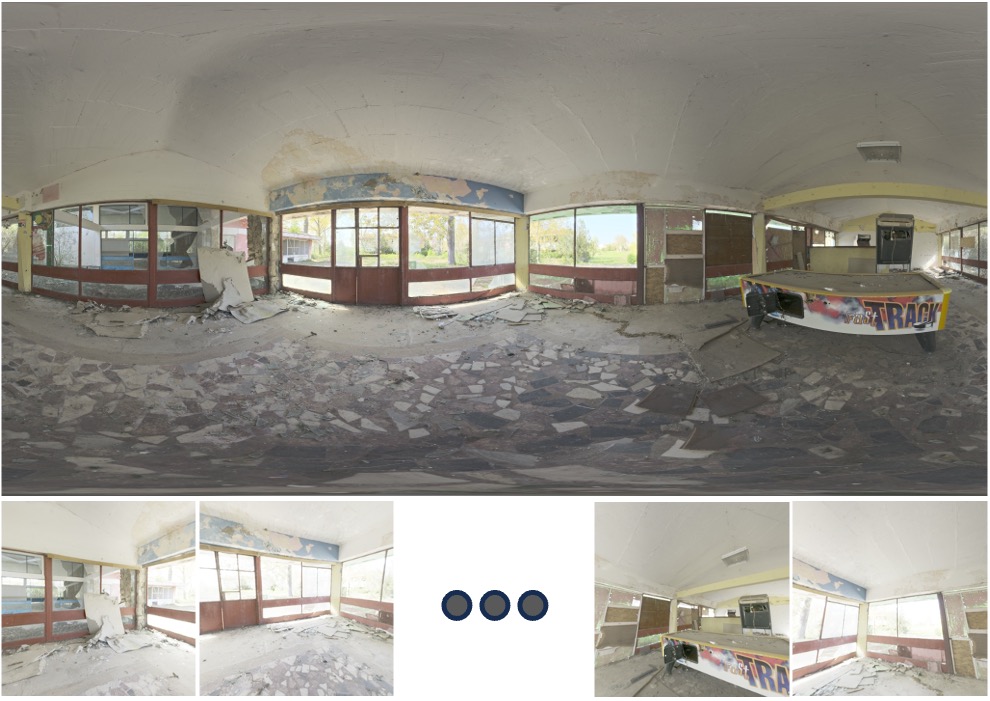}
    \caption{To augment the perspective HDRI dataset, we generate 20 perspective HDR images for every 360-degree equirectangular HDRI map. The images are tone mapped for visualization purpose.}
    \label{fig:rasterization}
\end{figure}

\paragraph{Text-to-image:} Most recent text-conditioned image generative models are based on diffusion~\cite{sohl2015deep, dhariwal2021diffusion, kingma2021variational}, a technique which generates new samples via progressive denoising from pure random noise. In order to reduce the computational footprint of such approaches, latent diffusion models, like Stable Diffusion (SD)~\cite{rombach2021high}, perform diffusion in the latent space of a variational autoencoder (VAE)~\cite{kingma2013auto}. Text-to-image generation methods have unlocked other use cases such as text-to-panorama with applications in virtual reality environments to enable immersive experiences. In particular, several works~\cite{bar2023multidiffusion, Tang2023mvdiffusion, skylab} have extended SD models to generate panorama from text. However, these approaches only consider horizontal (left-to-right) panoramic rotations. Lin et al.~\cite{lin2021infinitygan} use an adversarial setup for unconditioned seamless panoramic generation in a patch-wise manner. More recently, LDM3D-VR~\cite{stan2023ldm3d} fine-tunes a pre-trained SD v1.5 model to generate panoramic RGB images with monocular depth. However, the generated image contains a visible border seam due to mismatch in the generated content on both ends of the image. Most text-to-panoramic 360-degree approaches are limited in their applications due to the generation of LDR images. In contrast, HDRI panoramas can assist in synthesizing 360-degree photorealistic lighting and reflections for scene or portrait relighting. Text2Light~\cite{chen2022text2light} proposes a text-conditioned panoramic 360-degree HDRI map generation by using a complex dual-stage architecture. Although the generated images are in the HDRI space, the Text2Light framework lacks the image diversity compared to text-to-image models based on diffusion.  

\paragraph{Portrait relighting:}
The pioneering work by Debevec et al.~\cite{debevec2000acquiring} designs a spherical rig called Light Stage to capture a person’s reflectance fields as one-light-at-a-time (OLAT) images and uses image-based relighting to relight static faces. Other approaches \cite{wenger2005performance,guo2019relightables,meka2019deep} utilize time-multiplexed illumination or color gradient illumination to relight dynamic subjects, but these methods require expensive custom capture rigs that are known for being hard to set up.


With the advancement in mobile photography, several deep learning approaches to relight portrait images captured in unconstrained environments have emerged. Zhou et al. \cite{zhou2019deep} and Sun et al. \cite{sun2019single} are early works to apply deep learning to portrait relighting by employing an encoder-decoder network to take a single image as input, inject the target illumination into the bottleneck layer of the network, and output the re-illuminated image. More recently, pixel-aligned components such as normal, albedo, diffuse light map, specular light map, visibility map, and shadow map have been incorporated into network designs to improve performance \cite{nestmeyer2020learning, wang2020single, pandey2021total, hou2022face, song2022real, qiu2024relitalk}. 
However, these approaches often involve compute-heavy pipelines with multiple large networks, hindering the portability of these methods to mobile devices. Additionally, these approaches lack temporal consistency for in-the-wild videos.



In video relighting, it is critical to have a good balance between high-quality, video stability, and model complexity. Zhang et al. \cite{zhang2021neural} introduces the flow-based temporal loss supervised on dynamic OLAT dataset for explicit temporal modeling. Despite showing real-time relighting on mobile devices, the light-weight encoder-decoder network cannot produce high-quality relit videos with sufficient facial details. Yeh et al. \cite{yeh2022learning} propose to learn two temporal residual networks for improving consistency of intermediate normal and albedo predictions, but the network size is not designed for consumer device deployment.


Even several commercial relighting solutions do not support on-device inference and video relighting with HDRI environments. Portrait Mode on iPhone~\cite{apple} only provides $5$ studio lights mode, mostly for photo editing. Portrait Light on Google’s Pixel~\cite{google} and Google Meet~\cite{googlemeet} only assist in relighting a subject with point light sources, rather than using 360-degree environment light for illumination. Clipdrop~\cite{clipdrop} offers controllability of point light sources, but cannot relight videos. The recent SwitchLight~\cite{switchlight} supports video relighting, but it is a frame-based solution that run on a remote server, hence not a real-time solution. In contrast, our proposed video relighting method combines generalized normal estimation network and light adding based rendering, leveraging mobile computing to achieve convincing and coherent relighting results in real-time.

\section{Text-conditioned 360-degree HDRI map generation}

Our goal is to enable on-device real-time video portrait relighting by leveraging high dynamic range image (HDRI) maps from a text-to-image generative model for relighting with diverse background environments. To that end, we leverage the generative ability of Stable Diffusion~\cite{rombach2021high} (SD) which we extend to 360-degree HDRI map generation.





\subsection{Quantized HDR images}
\label{subsec:quantization}

To synthesize 360-degree images from user text prompts, we propose fine-tuning a pre-trained SD v1.5 model on HDRI panoramas. However, since our end goal is to achieve fast inference on 
the AI accelerator of a device with a Snapdragon\textsuperscript{\textregistered} Gen 3 Platform\footnote{\label{snapdragon}Snapdragon branded products are products of Qualcomm Technologies, Inc. and/or its subsidiaries.}, we aim to leverage 8-bit model quantization \cite{nagel2021white} at test time to reduce the computational and memory footprint of our text-to-image models, as previously done in \cite{qualcommblogpostsd}. Yet, it poses challenges for HDR prediction due to the disparity between the high-dynamic range input and output signals (with luminance range approximately 200,000 using FLOAT32), and the INT8 quantization applied to both weights and internal activations during inference. To mitigate this potential performance drop post-model quantization, we propose to preemptively quantize the raw FLOAT32 HDRI maps using the perceptual quantizer (PQ) based HDRI quantization workflow~\cite{standard2014high, series2012parameter} to obtain 8-bit images while preserving the high-dynamic luminance spectrum.



More specifically, we combine the perceptual quantizer (PQ) used in the HDR10 standard~\cite{standard2014high} with an 8-bit UINT quantization formula. As shown in Eq.~\ref{eq:pq_eotf_1}, first the PQ inverse electro-optical transfer function (EOTF) is used to convert from linear luminance to non-linear color values. Then, the quantization formula~\cite{series2012parameter} is applied to produce quantized HDR images in UINT8 domain:

\begin{align}
    E' &= PQ_{EOTF}^{-1}(F_D) \nonumber \\
    PQ_{EOTF}^{-1}(F_D) &= \Biggl(\frac{c_1 + c_2 \cdot Y^{m_1}}{1 + c_3 \cdot Y^{m_1}}\Biggr)^{m_2} \label{eq:pq_eotf_1} \\
    D' &= int(scale \cdot E') \nonumber
\end{align}
where $F_D$ is the linear luminance in $cd/m^{2}$, $E'$ is the non-linear color value, $Y=F_D/10000$, $m_1=0.1593017578125$, $m_2=78.84375$, $c_1=0.8359375$, $c_2=18.8515625$, and $c_3=18.6875$. For the quantization formula, we set $scale=198$ to encode luminance in original HDRI maps at a maximum of $200,000$ $cd/m^{2}$. The scale is set based on statistical analysis: among 624 real HDRI panoramas in PolyHaven\cite{polyhaven}, only 33 panoramas have an average $4$ pixels with values greater than $200,000$.

Post fine-tuning the SD v1.5~\cite{rombach2021high} on UINT8 quantized HDR images ($D'$), at inference, the generated quantized HDRI map is transformed to the original HDRI space by applying the dequantization followed by PQ EOTF as indicated in Eq.~\ref{eq:pq_eotf_2}. An overview of the images generated from the forward and reverse HDRI quantization are shown in Fig.~\ref{fig:pq_eotf}.

\begin{align}
    E' &= D'/scale \nonumber \\
    F_D &= PQ_{EOTF}(E') \label{eq:pq_eotf_2} \\
    PQ_{EOTF}(E') &= 10000 \Biggl(\frac{max[(E'^{1/m_2} - c_1), 0]}{c_2-c_3 
    \cdot E'^{1/m_2}}\Biggr)^{1/m_1} \nonumber
\end{align}

\begin{figure}[!t]
    \centering
    \includegraphics[width=0.475\textwidth]{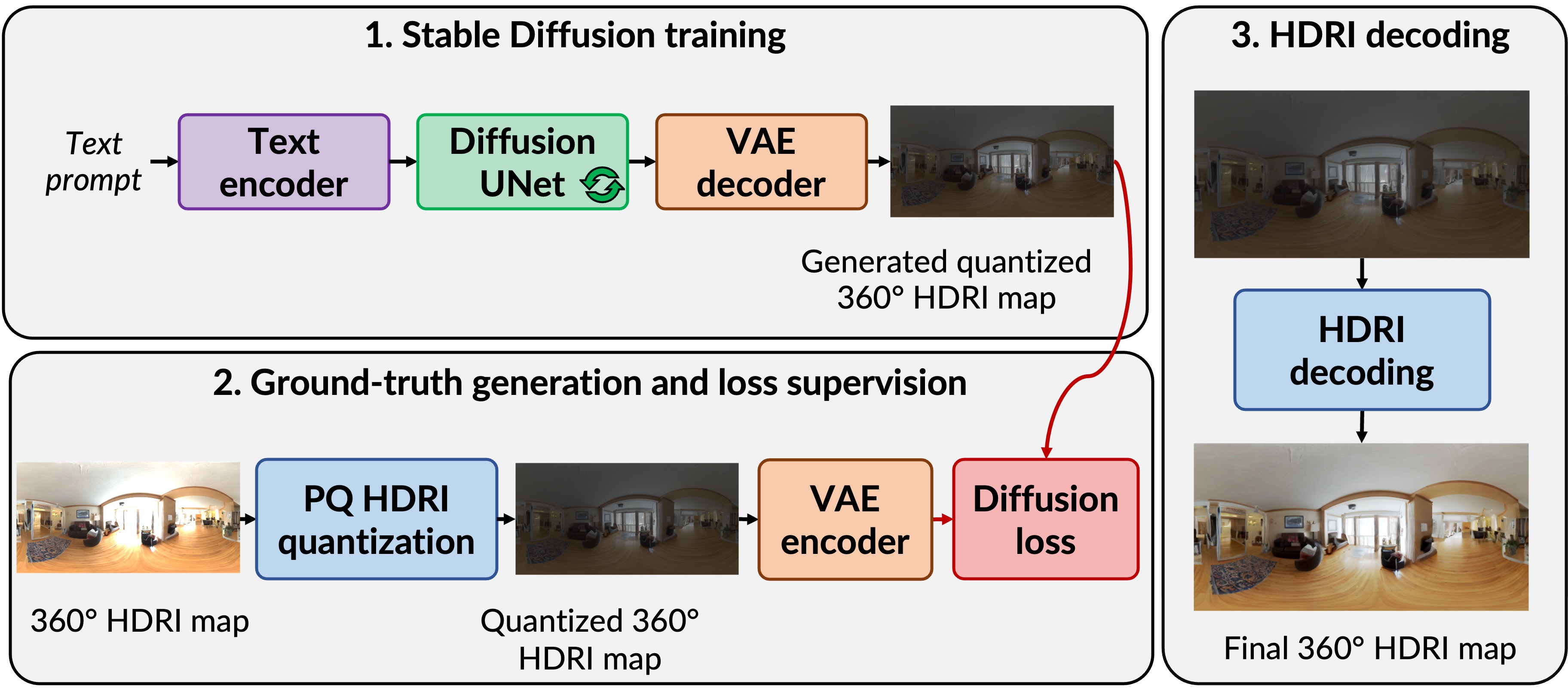}
    \caption{An overview of the (1) text to 360-degree training setup along with (2) the quantized HDRI generation. The generated quantized image can be (3) dequantized with inverse PQ transformation to produce the final HDRI map.}
    \label{fig:method_panorama}
\end{figure}

\begin{figure}[!t]
    \centering
    \includegraphics[width=0.475\textwidth]{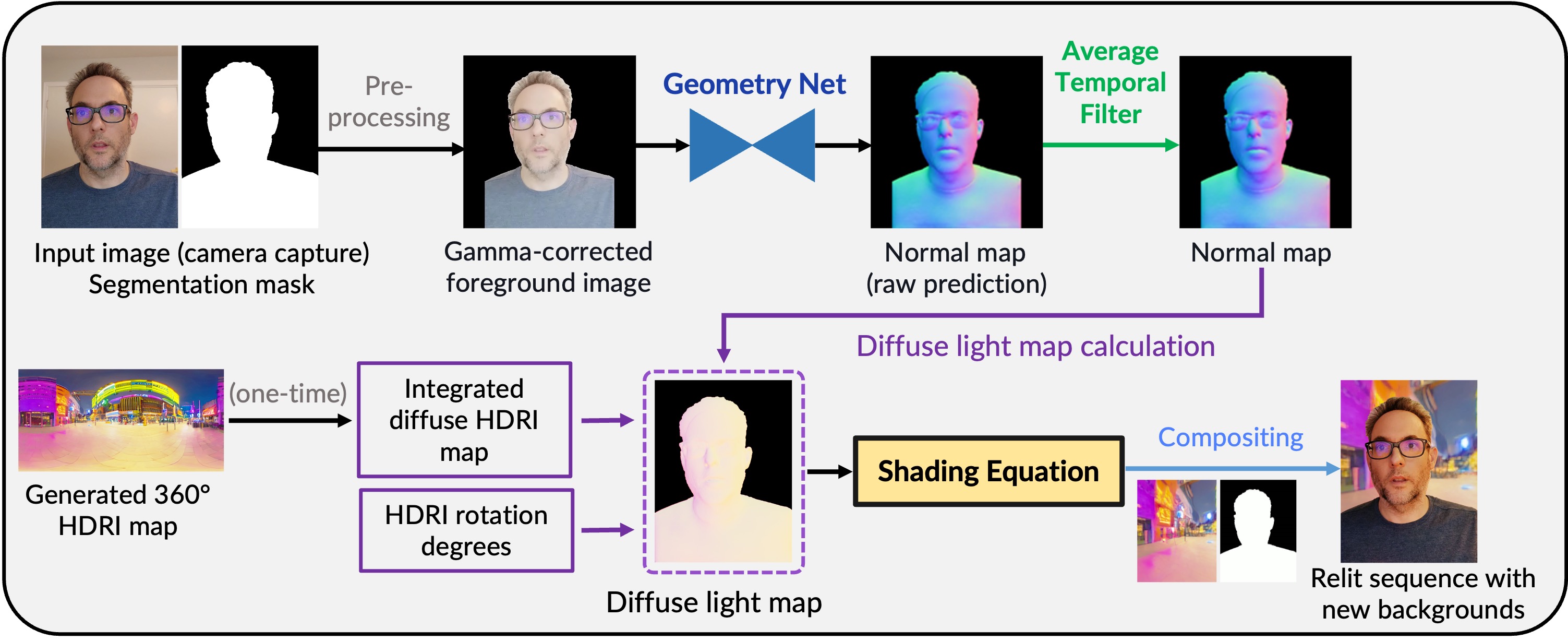}
    \caption{Overview of the proposed relighting pipeline. Given an input image and segmentation mask from the camera, we generate a gamma-corrected foreground image to obtain the surface normals from Geometry Net. The raw normal map prediction is temporally stabilized with an average temporal filter. To perform relighting, we first use the generated 360-degree HDRI map and the filtered normal map to compute the diffuse light map, and then use the shading equation to relight the portrait.}
    \label{fig:method_relighting}
\end{figure}

\subsection{Training setup}

\textbf{Training data}: We combine PolyHaven~\cite{polyhaven} and Laval~\cite{gardner2017learning, hold2019deep}, two publicly accessible yet relatively small-scale datasets offering 360-degree HDR images. PolyHaven contains around $660$ high-resolution equirectangular images depicting a variety of indoor and outdoor scenes. Laval, on the other hand, provides a collection of $2100$ indoor and $206$ outdoor 360-degree images. To augment our training data, we also utilize Hypersim~\cite{roberts2021hypersim}, which consists of standard HDR images, particularly during VAE fine-tuning. We apply the transfer function specified in Eq.~\ref{eq:pq_eotf_1} to all three datasets to generate quantized HDRI maps. 


\textbf{VAE}: Utilizing the original VAE from the SD v1.5~\cite{rombach2021high} pipeline to encode quantized HDR images leads to notable color distortions and artifacts in the decoded images, as shown in Fig.~\ref{fig:vae}.  
To address this, we fine-tune the VAE on quantized HDRI data. Given that the VAE does not require to specialize on 360-degree images, we can substantially expand our dataset for this phase. Rather than directly using the full 360-degree equirectangular images from our combined PolyHaven-Lavel dataset, we construct a data augmentation pipeline which entails rendering perspective views from these 360-degree images, as shown in Fig~\ref{fig:rasterization}. Moreover, to further expand our training dataset, we incorporate HDR images from the Hypersim~\cite{roberts2021hypersim} dataset. Post-training, as depicted in Fig.~\ref{fig:vae}, our fine-tuned VAE successfully reconstructs HDR images devoid of color distortions or other artifacts initially observed with the original model.

\textbf{Diffusion UNet}: To fine-tune the original SDv1.5 diffusion U-Net for text-conditioned 360-degree HDR generation, we adopt a two-stage training strategy to address the limited availability of 360-degree HDRI data. In the first stage, the objective is to adapt the diffusion U-Net to the fine-tuned latents and learn a prior on the luminance range (HDR domain adaptation). For this, we do not use 360-degree images and instead reuse the augmented, perspective-based, quantized HDRI dataset constructed during VAE fine-tuning.
Text prompts are generated using the BLIP~\cite{li2022blip} image captioning model, applied to tone mapped low dynamic range (LDR) versions of the images.

In the second stage, we further fine-tune the latent diffusion U-Net on 360-degree images this time. We use the quantized images from PolyHaven and Laval along with their BLIP-based captions. Remarkably, even with a relatively limited dataset comprising only a few thousand 360-degree images, the model effectively learns the capability to generate consistently accurate 360-degree imagery. During inference, we revert the generated quantized images back to the original HDRI format using the inverse PQ function outlined in Eq.~\ref{eq:pq_eotf_2}, thereby readying them for downstream tasks such as portrait relighting. To ensure 360-degree consistency at the edges of the generated equirectangular image, we adopt the circular latent padding technique proposed in~\cite{wang2023360} throughout all denoising timesteps. An overview of the training, quantized data preparation and inference is illustrated in Fig.~\ref{fig:method_panorama}.

\section{Video portrait relighting}



We propose a relighting pipeline to generate relit portrait videos with temporal stability, high-quality and with light-weight network architecture. Our proposed setup involves a Geometry Net for surface normal map prediction, a average temporal filter for enhancing temporal consistency, diffuse light map calculation, a shading equation for rendering relit foreground, and background composition as shown in Fig.~\ref{fig:method_relighting}.

\begin{table}[!t]
    \centering
    \begin{adjustbox}{}
        \small
        {\renewcommand{\arraystretch}{1.1}
            \begin{tabular}{l|c|c}
            \hline
            \textbf{Method} & \multicolumn{1}{c|}{\textbf{Runtime}} & \multicolumn{1}{c}{\textbf{Memory}}\\
            \multicolumn{1}{c|}{} & {(seconds)} & {(GB)} \\ 
            \hline
            
            Text2Light~\cite{chen2022text2light} (2048$\times$4096) & 61.25 & 3.1
                        \\ \hline
            Ours (512$\times$512)& 5.2& $\sim$3.4 \\
            Ours (512$\times$1024)& 14.8& $\sim$3.4\\

            \hline
            \end{tabular}
        }
    \end{adjustbox}
    \caption{We measure the runtime and memory requirements of Text2Light~\cite{chen2022text2light} and our proposed 360-degree generative approach on a single NVIDIA A100 GPU as Text2Light does not support on-device inference.}
    \label{tab:runtime_gpu}
\end{table}
\begin{table*}[!t]
    \centering
    \begin{adjustbox}{}
        \scriptsize
        {\renewcommand{\arraystretch}{1.1}
            \begin{tabular}{l|c|c|c|c|c|c|c}
            \hline
            \textbf{Method} & \multicolumn{1}{c|}{\textbf{Image}} & \multicolumn{1}{c|}{\textbf{Video}} & \multicolumn{1}{c|}{\textbf{Run on}} & \multicolumn{1}{c|}{\textbf{Video}} & \multicolumn{1}{c|}{\textbf{HQ}} & \multicolumn{1}{c|}{\textbf{Target}} & \multicolumn{1}{c}{\textbf{Controllability}}\\

            \textbf{} & \multicolumn{1}{c|}{\textbf{relighting}} & \multicolumn{1}{c|}{\textbf{relighting}} & \multicolumn{1}{c|}{\textbf{mobile device}} & \multicolumn{1}{c|}{\textbf{consistency}} & \multicolumn{1}{c|}{\textbf{relighting}} & \multicolumn{1}{c|}{\textbf{light}} & \multicolumn{1}{c}{\textbf{}}\\
            
            \hline
            
            Apple iPhone~\cite{apple}
                & \greencheck 
                & \redcross 
                & \redcross 
                & \redcross
                & \greencheck 
                & Studio light
                & Limited lighting options and no env. rotation
                \\
            Google Pixel~\cite{google}
                & \greencheck 
                & \redcross 
                & \redcross 
                & \redcross
                & \greencheck 
                & Single point light
                & Light direction and intensity
                \\
             Google Meet~\cite{googlemeet}
                & \greencheck 
                & \greencheck 
                & \redcross 
                & \greencheck
                & \greencheck 
                & Multiple point lights
                & Light direction, color and intensity
                \\
            Clipdrop~\cite{clipdrop} 
                & \greencheck 
                & \redcross 
                & \redcross 
                & \redcross
                & \greencheck 
                & Multiple point lights
                & Light direction, intensity, color, distance and radius
                \\
            SwitchLight~\cite{switchlight}
                & \greencheck 
                & \greencheck 
                & \redcross
                & \redcross
                & \greencheck 
                & 360$^{\circ}$ HDR image
                & HDR rotation degrees
                \\
            Zhang et al.~\cite{zhang2021neural}
                & \greencheck 
                & \greencheck
                & \greencheck
                & \greencheck
                & \redcross 
               & 360$^{\circ}$ HDR image
                & HDR rotation degrees
                \\
            Yeh et al.~\cite{yeh2022learning}
                & \greencheck 
                & \greencheck 
                & \redcross 
                & \greencheck
                & \greencheck 
                & 360$^{\circ}$ HDR image
                & HDR rotation degrees
                \\ \hline
            \textbf{Ours} 
                & \greencheck 
                & \greencheck 
                & \greencheck 
                & \greencheck 
                & \greencheck 
                & 360$^{\circ}$ HDR image
                & HDR rotation degrees
                \\
        
            \hline
            \end{tabular}
        }
    \end{adjustbox}
    \caption{Unlike other existing approaches, our proposed relighting method supports all key features such as image and video relighting, on-device inference, video consistency, high-quality relighting, relight with 360-degree HDR environment maps, and fine-grained control over the HDR rotation.}
    \label{tab:relighting_comparisons}
\end{table*}
\begin{table}[!t]
    \centering
    \begin{adjustbox}{}
        \small
        {\renewcommand{\arraystretch}{1.1}
            \begin{tabular}{l|c|c}
            \hline
            \textbf{Method} & \multicolumn{1}{c|}{\textbf{Runtime}} & \multicolumn{1}{c}{\textbf{Model Size}}\\
            & \multicolumn{1}{c|}{(seconds)} & \multicolumn{1}{c}{(MB)}\\
            \hline
            
            Text-to-360-degree image & $\sim 5$ & $\sim 1100 $
                        \\ \hline
            Video Relighting & $\sim 0.04$ & $\sim 20.5$ \\
                        
            \hline
            \end{tabular}
        }
    \end{adjustbox}
    \caption{Overview of the on-device runtime and model size for both the 360-degree image generation and video relighting pipelines on a Snapdragon Gen 3 platform. Note that the model size for video relighting is the sum of the size of the video segmentation network ($\sim 17.3$ MB) and the Geometry Net ($\sim 3.2$ MB).}
    \label{tab:runtime}
\end{table}


\begin{figure}[!t]
    \centering
    \includegraphics[width=0.475\textwidth]{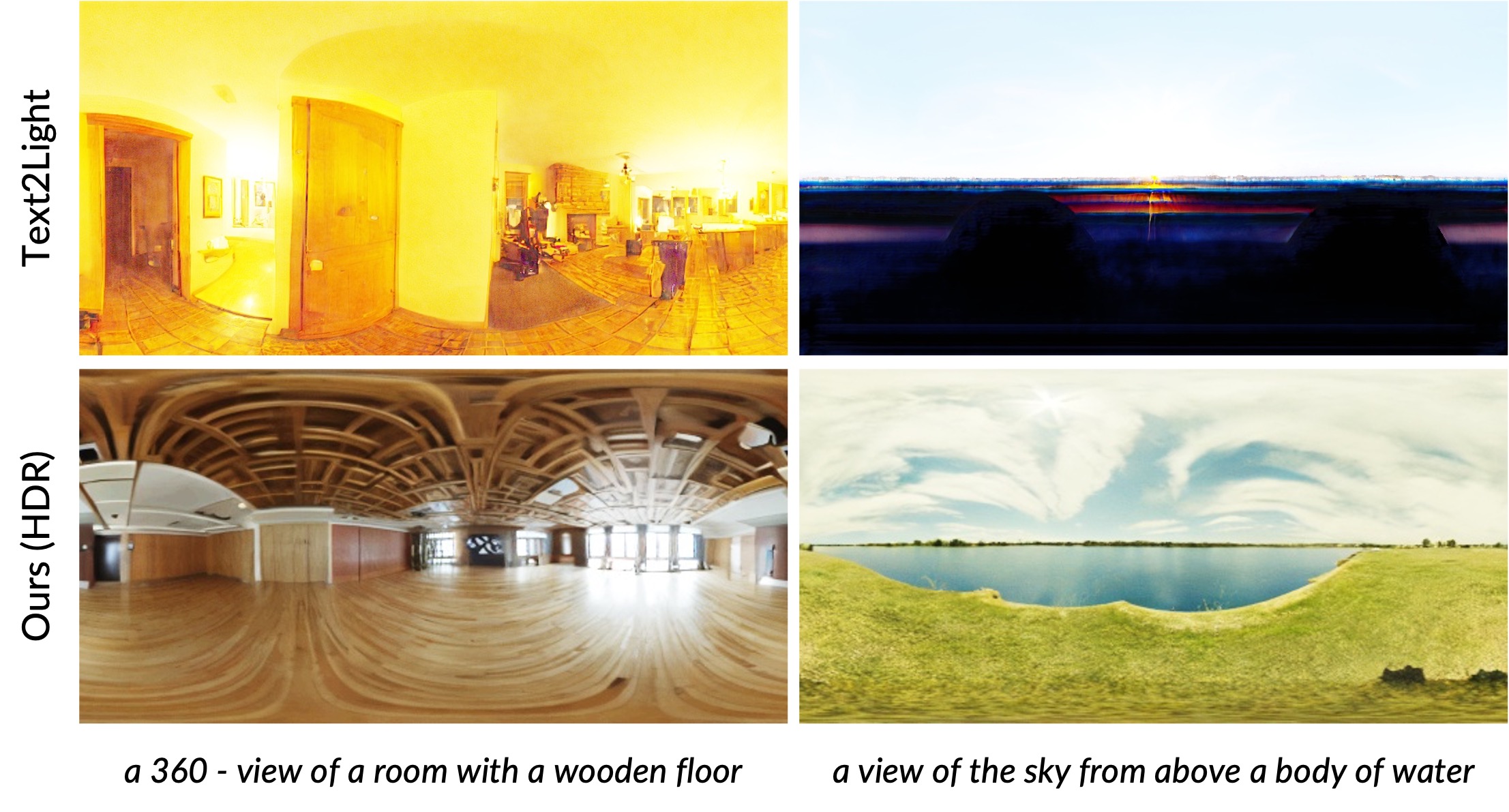}
    \caption{Our Stable Diffusion~\cite{rombach2021high} based text-conditioned 360-degree HDRI generative model can generate realistic and diverse environments compared to existing Text2Light~\cite{chen2022text2light} model. The images are tone mapped for visualization purpose.}
    \label{fig:stable360hdr_comparison}
\end{figure}

\begin{figure*}[!t]
    \centering
    \includegraphics[width=0.9\textwidth]{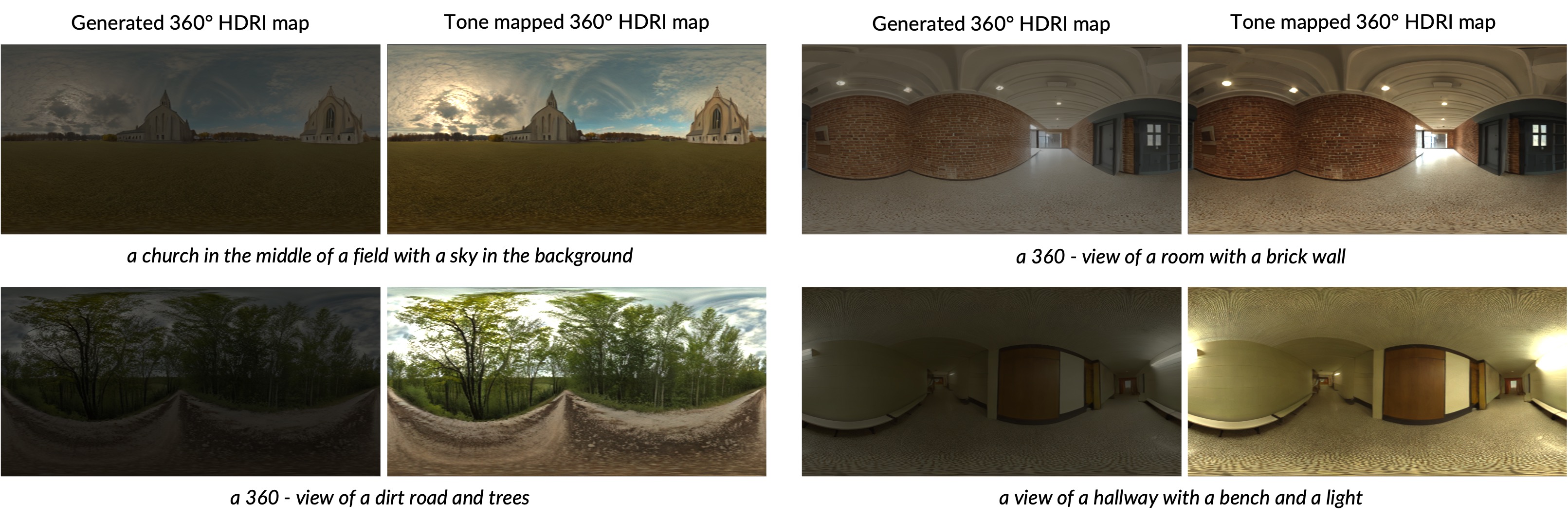}
    \caption{Our text to 360-degree HDRI generative model can produce environment maps for diverse text prompts. We show both the generated quantized HDR image and it's dequantized with inverse PQ, and tone mapped image.}
    \label{fig:stable360hdr}
\end{figure*}

\subsection{Geometry Net}

To estimate the surface normals, inspired by the Total Relighting~\cite{pandey2021total} framework, we develop a light-weight Geometry Net with a similar but smaller UNet architecture with 13 layers. In addition, the size of network input is set to $512 \times 512$ to reduce computational complexity and to handle both portrait and landscape camera captures. The Geometry Net aims to produce per-pixel surface normal map $N_{t}$ as a geometrical cue to calculate physically correct diffuse light map. Given an input camera capture $I$, we leverage our proprietary segmentation network with on-device support to extract the foreground segmentation, but we can leverage any off-the-shelf segmentation network for this task. To prepare the input to the Geometry Net, the camera capture $I$ is first gamma corrected and masked using the foreground segmentation, then resized and padded to $512 \times 512$. As for training, due to the challenges of setting up a light stage, inspired by~\cite{wang2020single}, we use head meshes captured by the 3DMD system~\cite{3dmd}, HDRI panoramas in PolyHaven\cite{polyhaven}, and 3D software Blender~\cite{blender} to create the synthetic normal dataset. The synthetic normal dataset contains paired rendered images and normal ground truths in camera space coordinates for 60 identities, each with 31 expressions. Please refer to the supplementary material for more details on the network architecture and the creation of synthetic normal dataset.


\subsection{Average Temporal Filter}
When applying the Geometry Net to video sequences, minor flickering occurs between consecutive normal estimations because the network is frame-based. In addition, since the head meshes in our synthetic normal dataset lack hair and cloth geometry, the instability issue around hair and cloth areas are severer. To further improve the temporal consistency while maintaining low computation complexity at edge, we do not add an additional temporal refinement network to the architecture as~\cite{yeh2022learning}, but adopt an effective solution to apply the average temporal filter on three consecutive normal maps $N_{t-2}$, $N_{t-1}$, and $N_{t}$ predicted by the Geometry Net. The operation is formulated in Eq.~\ref{eq:avg_filter}:
\begin{equation}
    \widetilde{N_t} = \frac{1}{3}(N_{t-2} + N_{t-1} + N_t)
    \label{eq:avg_filter}
\end{equation}
where $\widetilde{N_t}$ is the average normal map used to calculate the diffuse light map, which helps in mitigating flickering issues in final relit sequences.

\begin{figure}[!t]
    \centering
    \includegraphics[width=0.475\textwidth]{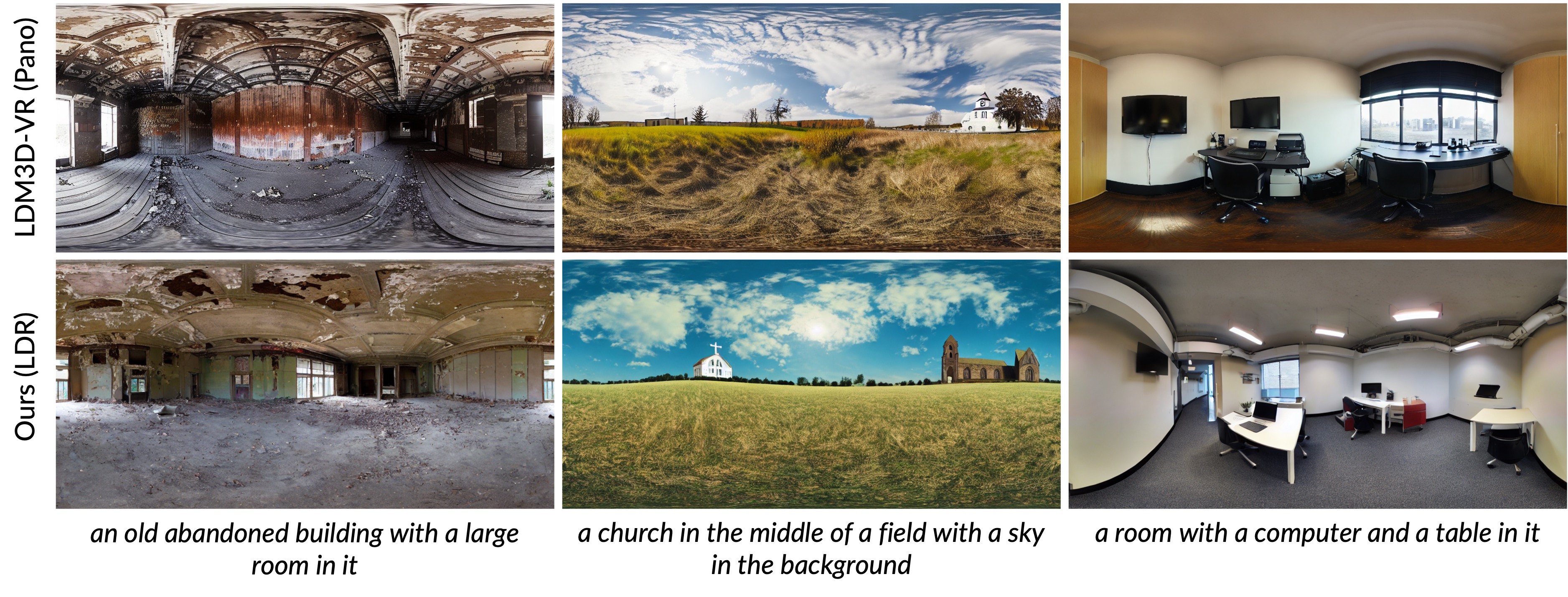}
    \caption{To demonstrate the strong generative capabilities of our proposed text to 360-degree image generation, we compare the results from an LDR variant of the model with LDM3D-VR~\cite{stan2023ldm3d}. For more qualitative comparison, please refer to the supplementary material.}
    \label{fig:stable360ldr}
\end{figure}



\subsection{Light Adding based Rendering}

Unlike state-of-the-art relighting methods~\cite{pandey2021total,yeh2022learning} which require at least four sequential networks, to save calculations, we explore combining a neural network with physically based rendering (PBR). Additionally, when the subjects are illuminated with physically reasonable diffuse light, we observe that they appear to be in the virtual environment. Therefore, we propose a light adding based rendering to produce a relit foreground by adding diffuse shading to the camera capture. Specifically, given a generated 360-degree HDRI map, we pre-compute the diffuse cubemap by summing all diffuse light from the surrounding environment along each sample direction. For each frame, a diffuse light map can be calculated by sampling the pre-integrated diffuse cubemap using per-pixel normal vectors, which is a popular technique in real-time PBR implementation~\cite{pbrIBL_diffuse}. To further render relit foregrounds, the shading equation Eq.~\ref{eq:relight_1} is formulated as a scaled addition of raw camera capture and diffuse shading, expressed as the multiplication of the low saturation camera capture $I_{lowS}$ and the diffuse light map $D$:

\begin{equation}
    R = s_1 \cdot I + s_2 \cdot I_{lowS} \odot D    
    \label{eq:relight_1}
\end{equation}

where $\odot$ denotes the element-wise multiplication, and $s_1$,$s_2$ are scaling constants. Note that generally the diffuse shading is formulated as the multiplication of albedo and diffuse light map~\cite{nestmeyer2020learning,pandey2021total, pbrIBL_diffuse}. Considering that adding an sequential  network after Geometry Net increases the operation time by $\sim24$ milliseconds for a $1024\times768$ albedo estimation, it introduces a computational bottleneck for on-device real-time video relighting. Therefore, we compute a low saturation camera capture $I_{lowS}$ using Eq.~\ref{eq:relight_lowS} instead:
\begin{equation}
    I_{lowS} = 0.6 \cdot I + 0.4 \cdot I_{gray} + 0.05
    \label{eq:relight_lowS}
\end{equation}

where $I_{gray}$ is the grayscale camera capture. Specifically, $I_{lowS}$ is designed to be computationally efficient without requiring color space conversion from RGB to HSV, and the low saturation is designed to reduce the impact of the input lighting’s hue.

\begin{figure}[!t]
    \centering
    \includegraphics[width=0.475\textwidth]{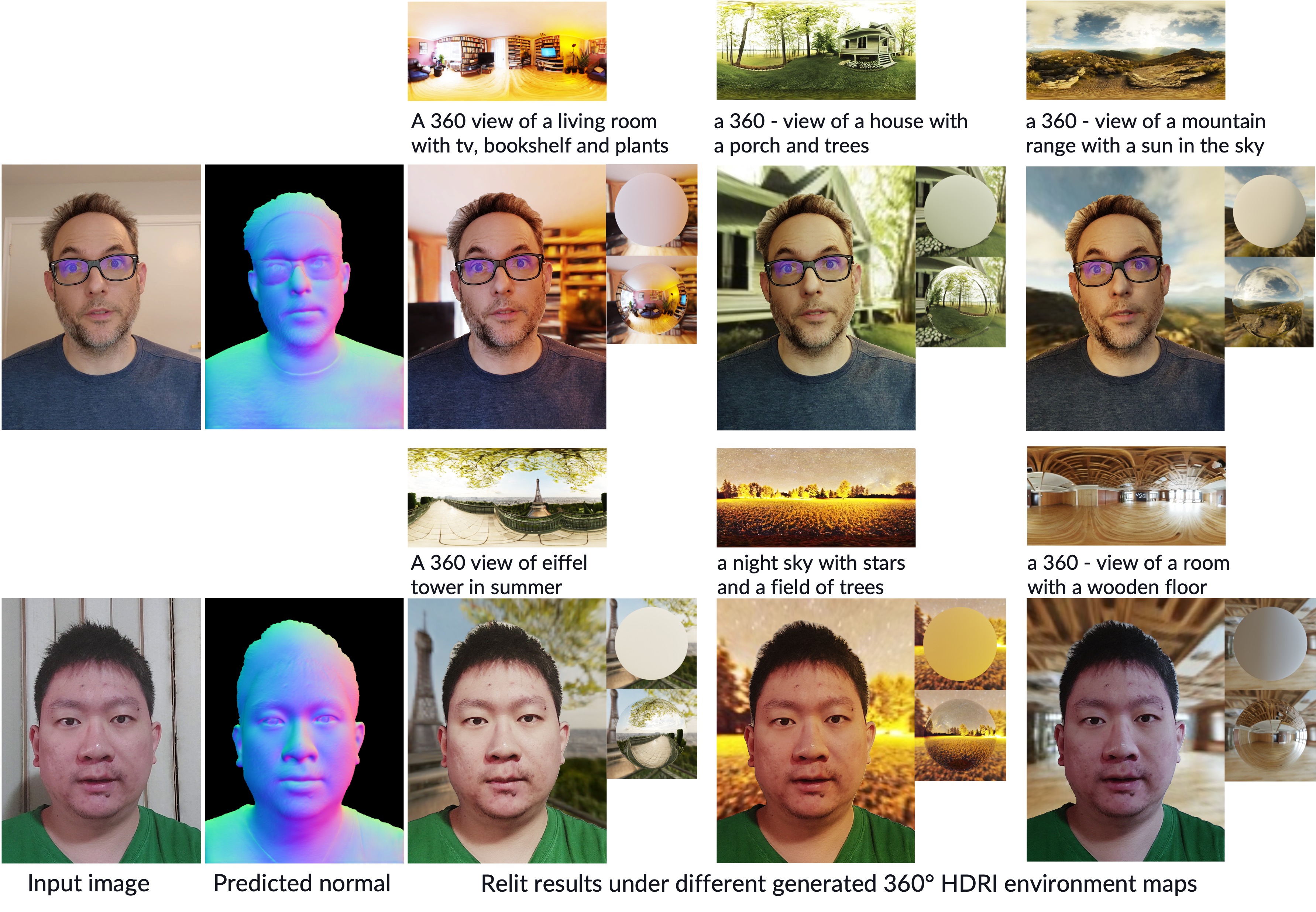}
    \caption{Our proposed framework shows realistic relighting of different portraits on diverse generative HDRI environment maps. Our results can preserve facial details such as wrinkles and beards and generate lighting effects that are consistent with diffuse sphere references.
    }
    \label{fig:different_portraits}
\end{figure}



\section{On-device inference}
To enable inference on Snapdragon Gen 3 platform, we leverage network quantization and real-time rendering. For on-device network inference, we use the AIMET~\cite{siddegowda2022neural} library to conduct post-training quantization from FP32 to INT8 for both the 360-degree image generation and video relighting models. Additionally, we implement the light adding based rendering module in OpenGL Shading Language~\cite{opengl} to save expensive floating-point operations on the CPU.

\begin{table}[!t]
    \centering
    \begin{adjustbox}{}
        \small
        {\renewcommand{\arraystretch}{1.1}
            \begin{tabular}{l|c|c}
            \hline
            \textbf{Method} & \multicolumn{1}{c|}{\textbf{CLIP $\uparrow$}} & \multicolumn{1}{c}{\textbf{FID $\downarrow$}}\\
            \hline
            
            LDM3D-VR~\cite{stan2023ldm3d} &  28.73 & 42.04
                        \\
            Ours-LDR & \textbf{29.94} & \textbf{39.79}
                        \\
                        
            \hline
            \end{tabular}
        }
    \end{adjustbox}
    \caption{We measure the CLIP and FID metrics on the LDR images generated by LDM3D-VR and our LDR 360-degree model to demonstrate the better generative capabilities.}
    \label{tab:stable360_ldr}
\end{table}

\section{Results}

Our proposed framework achieves real-time video portrait relighting based on text-prompted 360 degree image generations. We now present quantitative and qualitative analyses to showcase the capabilities of our proposed pipeline. 


\subsection{Text to 360-degree HDRI map generation} 

We compare images generated by our text-conditioned 360-degree HDRI generation against the Text2Light~\cite{chen2022text2light} approach in Fig.~\ref{fig:stable360hdr_comparison} and report the corresponding runtime on a single A100 GPU for both approaches in Tab.~\ref{tab:runtime_gpu}. Although the Text2Light model is capable of generating high-resolution 360-degree HDRI maps from text, it exhibits a lack of diversity and realism while requiring a significantly higher runtime. Our approach, on the other hand, can generate high-quality 360-degree images in $\sim5$ seconds on an A100 . More text-conditioned HDRI generations for both indoor and outdoor scenes are shown in Fig.~\ref{fig:stable360hdr}. Note that, for the sake of visualization, we show the corresponding tone mapped images.



We also conduct a study to compare with the recent LDM3D-VR~\cite{stan2023ldm3d} method, which proposes a similar approach to ours but is limited to low-dynamic range prediction. For this study, we adapt our training setup and fine-tune the SDv 1.5 U-Net on 360-degree LDR images from PolyHaven~\cite{polyhaven} and Matterport360~\cite{bata1126} with an image resolution of $512 \times 1024$ to match LDM3D-VR. We report FID~\cite{heusel2017gans} and CLIP~\cite{hessel2021clipscore} scores for both models in Tab.~\ref{tab:stable360_ldr}. On both metrics, our fine-tuned SD v1.5 model shows better performance. This can be attributed to (i) our model's capacity to generate consistent 360-degree image thanks to circular padding~\cite{wang2023360}, and (ii) the fact that we only use real-world images for fine-tuning unlike LDM3D-VR. Furthermore, Fig.~\ref{fig:stable360ldr} provides a qualitative comparison to demonstrate the superior quality of our generated images. For an extended qualitative analysis of the generated LDR images, we refer readers to the supplementary materials.

\subsection{Video portrait relighting}

To evaluate the proposed video portrait relighting method, first, we relight in-the-wild portrait sequences under different HDRI environments generated by our 360-degree image generative model. Due to the lack of relight ground truth for in-the-wild sequence, we render diffuse spheres to provide a lighting reference in the target HDRI environment. And mirror spheres are also provided to indicate where the main light source locates. As shown in Fig.~\ref{fig:different_portraits}, we are able to produce physically-correct relighting results while preserve facial details and can embed subjects naturally into a variety of generated environments. Additionally, to show temporal-consistency, we provide the recorded screenshots of our on-device EdgeRelight360 application in the supplementary material, demonstrating that our approach is stable and flicker-free.

\begin{figure}[!t]
    \centering
    \includegraphics[width=0.475\textwidth]{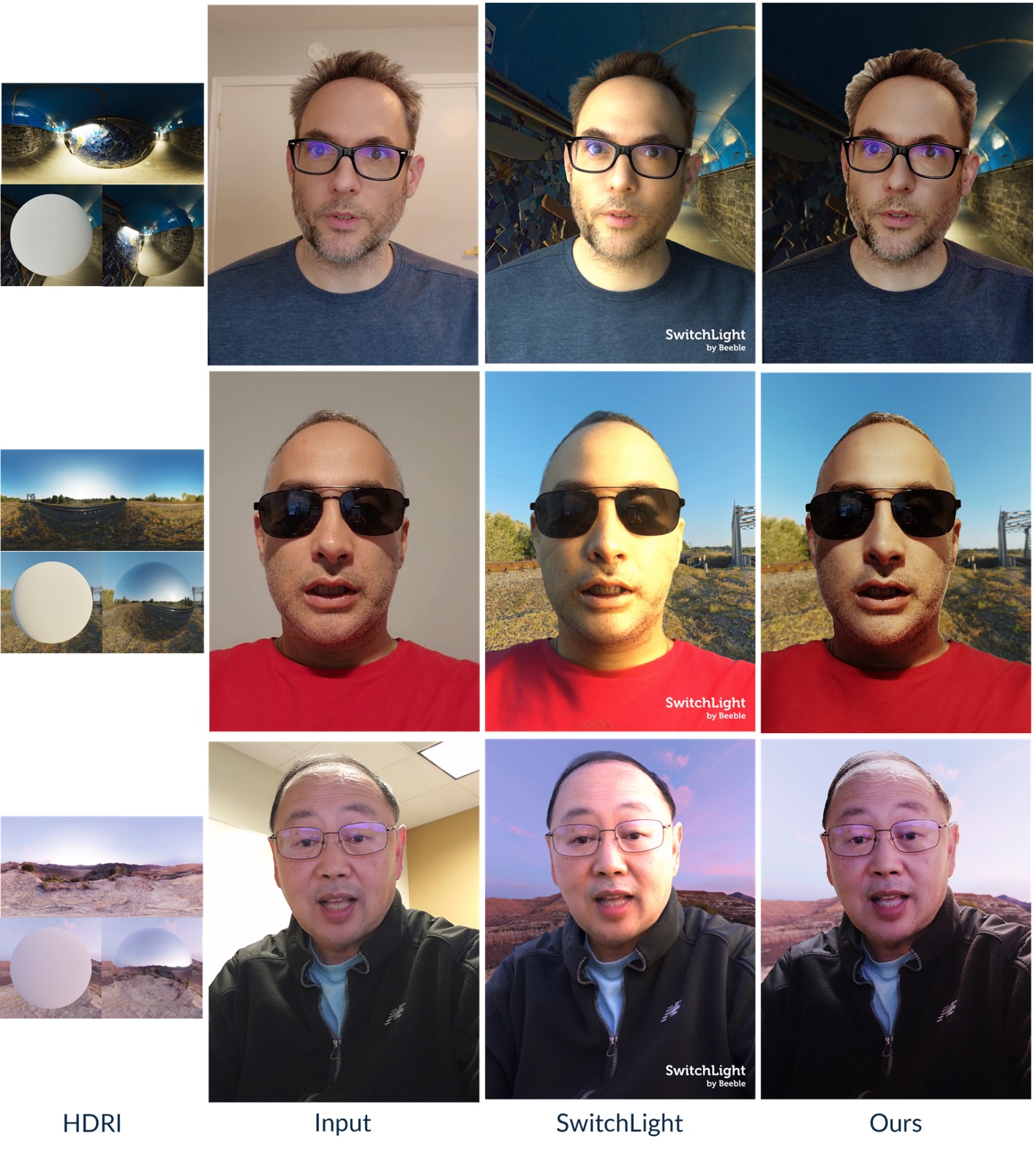}
    \caption{Compared with SwitchLight~\cite{switchlight}, our relighting pipeline is more lightweight and can generate more natural, physically correct, and temporally consistent results.}
\label{fig:relighting_switchlight}
\end{figure}

Second, we compare the runtime and relighting quality of publicly available SwitchLight~\cite{switchlight}, which also use HDRI maps to relight images. The web version of SwitchLight takes $10$ more seconds to run a single image on their remote server, while our method runs locally on the phone and takes only $0.04$ seconds per image. Their high computation cost origins from a complete intrinsic decomposition pipeline (including normal, albedo, specular, and roughness estimation), a neural renderer, and fine-grained foreground matting. Fig.~\ref{fig:relighting_switchlight} shows that we can produce promising results and make subjects blend into the PolyHaven~\cite{polyhaven} HDRI map naturally, while the default lighting effects of SwitchLight is too strong and less consistent with the lighting reference. Also, consecutive frames from our talking test sequence is put in the supplementary material to demonstrate that results from the web version of SwitchLight produce relit images with flickering, while ours are more stable and flicker-free. 

\subsection{On-device inference}
The primary goal of our proposed approach is to enable end-to-end on-device inference. To the best of our knowledge, we are the first to run both 360-degree HDRI map generation and real-time video portrait relighting. We compare some prior approaches that address image/video relighting with and without on-device support in Tab.~\ref{tab:relighting_comparisons}. In comparison to all the approaches, our proposed framework supports: (a) image and video relighting, (b) runs on device, (c) ensure smooth temporal consistency, (d) enables high-quality relighting, (e) leverages 360-degree HDRI maps, and (f) offers fine-grained rotation control of the HDRI maps. In contrast, all the prior approaches only handle a subset of the features.

In Tab.~\ref{tab:runtime} we show the runtime and model size of our proposed framework deployed on a mobile device with Snapdragon Gen 3 processor. The time to generate a single 360-degree HDRI map is $\sim 5$ seconds. For the end-to-end process which performs face detection, video segmentation, and video relighting concurrently, it runs at around $25$ fps using the neural signal processor (NSP) and GPU. Note that we generate the 360-degree images at $480 \times 480$ resolution with 20 denoising steps and run the Geometry Net at $512 \times 512$, while running rendering and produce relighting results at $1024 \times 768$.

\section{Discussion}

EdgeRelight360 supports real-time video applications by introducing an innovative, on-device video portrait relighting technique. By harnessing the power of text-conditioned generated 360-degree HDRI maps, it offers high-quality, realistic lighting conditions derived from textual descriptions. This not only ensures privacy and low runtime but also provides an immediate response to changes in lighting conditions or user input. A potential improvement to the existing 360-degree generation is to generate higher resolution images to support different edge screen resolutions with high-quality background images.
{
    \small
    \bibliographystyle{ieeenat_fullname}
    \bibliography{main}
}

\clearpage
\setcounter{page}{1}
\maketitlesupplementary

\paragraph{Geometry Net} 
Inspired from Total Relighting~\cite{pandey2021total}, we adopt a similar but more lightweight UNet architecture. It contains total 13 layers, where each layer consists of a $3 \times 3$ convolution, batch normalization~\cite{ioffe2015batch}, and parametric ReLU~\cite{agarap2018deep}. The encoder layers contain [$16$, $32$, $64$, $64$, $128$, $256$] filters, $256$ for the bottleneck layer, and [$256$, $128$, $64$, $64$, $32$, $16$] for the decoder layers. The encoder uses max-pooling for down-sampling, while the decoder has bilinear up-sampling layers. We use the $L_1$ loss between ground truth normal images in the synthetic dataset and predicted normal maps for supervision. The parameter number of Geometry Net is $2.864M$ and MACs is $7.417G$ given a $512 \times 512$ RGB input.

\begin{figure}
    \centering
    \includegraphics[width=0.4\textwidth]{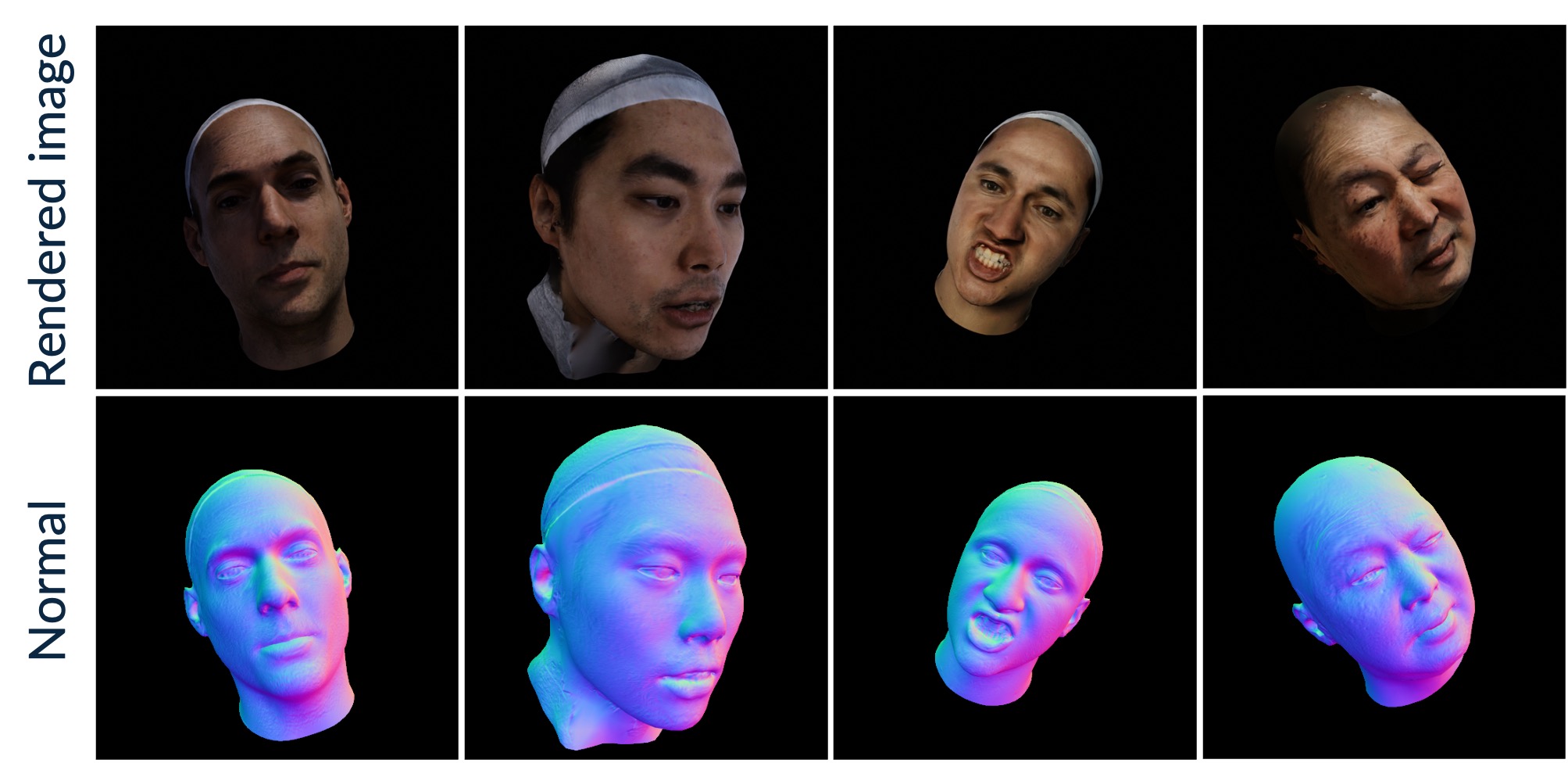}
    \caption{Examples in our synthetic normal dataset. Paired rendered images and view-dependent normal ground truths are rendered in Blender~\cite{blender} given 3D head meshes and HDRI environment maps.}
    \label{fig:normals}
\end{figure}

\paragraph{Synthetic Normal Dataset}
Inspired by~\cite{wang2020single}, first, we use the 3DMD system~\cite{3dmd} to capture head meshes of 60 subjects with 31 expressions and use the Wrap4D software~\cite{wrap4d} to automatically re-topologize them. Then, we use the ray-trace based render engine Cycles~\cite{blender_cycles} in Blender~\cite{blender} to render relit images and normal maps of head meshes under different environment lighting from PolyHaven~\cite{polyhaven}. To generate photorealistic renderings, we utilize tangent normals generated by the tool NormalmapGenerator~\cite{normalMapGenerator} to add surface details to the re-topologized head meshes. In addition, different from~\cite{wang2020single} which assumes a uniform specular coefficient $0.6$ during rendering, we use NormalmapGenerator~\cite{normalMapGenerator} to synthesize reasonable specular maps, indicating that specular reflections in areas such as forehead, cheeks, and nose are stronger.

In our synthetic normal dataset, among 60 identities, 53 identities are used for training and 7 identities for testing. And for the environment lighting in Blender, we randomly select 553 HDRI environments from PolyHaven~\cite{polyhaven} for training and 71 for testing. To add more variations in the synthetic dataset, the horizontal and vertical rotation angles of the HDRI environments are randomly set within [$0^\circ,  360^\circ$]  and [$-60^\circ,  60^\circ$]. The pitch, yaw, roll rotations of the head meshes are sampled in the range of [$-15^\circ,  15^\circ$], [$-45^\circ,  45^\circ$], and [$-30^\circ,  30^\circ$]. The y-axis position of the head mesh is randomly set within [$-0.4,  0.4$]. Overall, we render 300K paired samples and examples of our synthetic normal dataset are shown in Fig.~\ref{fig:normals}.

\begin{figure*}[!t]
    \centering
    \includegraphics[width=\textwidth]{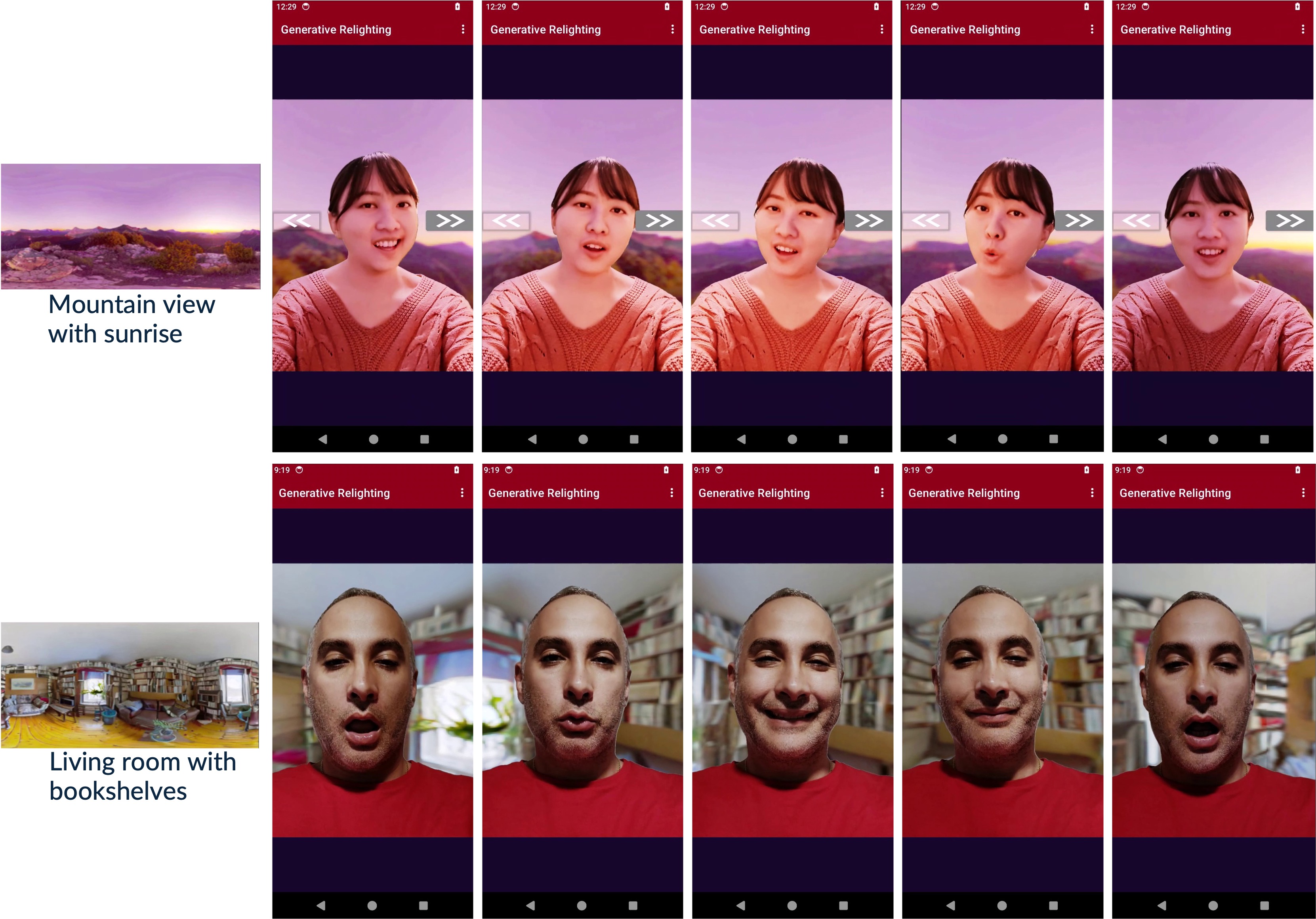}
    \caption{Our EdgeRelight360 application running on mobile device can generate temporally-consistent and realistic relit results on real talking sequences in real-time.}
    \label{fig:relighting_demo}
\end{figure*}

\begin{figure*}[!t]
    \centering
    \includegraphics[width=\textwidth]{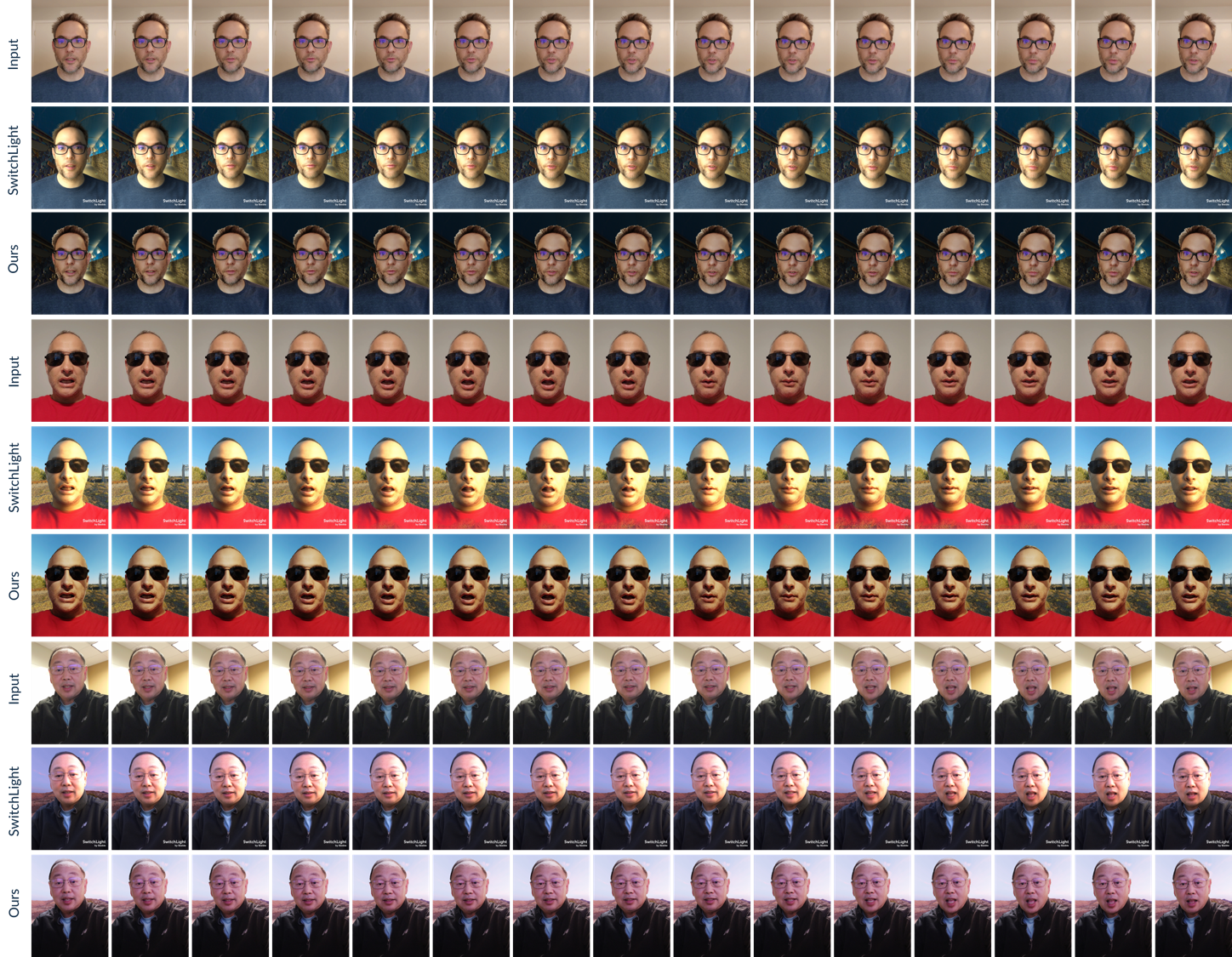}
    \caption{Our video relighting framework can generate more temporally-consistent relit results compared with the web version of SwitchLight\cite{switchlight}.}
    \label{fig:relighting_switchlight_temporal}
\end{figure*}

\paragraph{On-device Video Relighting}
In Fig.~\ref{fig:relighting_demo}, we show screenshots of on-device video relighting of talking sequences under different generative HDRI environments and rotations. Our supplementary video results exhibits good temporal consistency and very few flickering. Note that the scaling constants $s_1$ and $s_2$ in the shading equation are empirically set to $0.29$ and $0.38$ for all experiments with generative HDRI environments. Additionally, in Fig.~\ref{fig:relighting_switchlight_temporal}, we demonstrate the temporal consistency of our method and show the comparison with the web version of SwitchLight~\cite{switchlight}.



\begin{figure*}[!t]
    \centering
    \includegraphics[width=\textwidth]{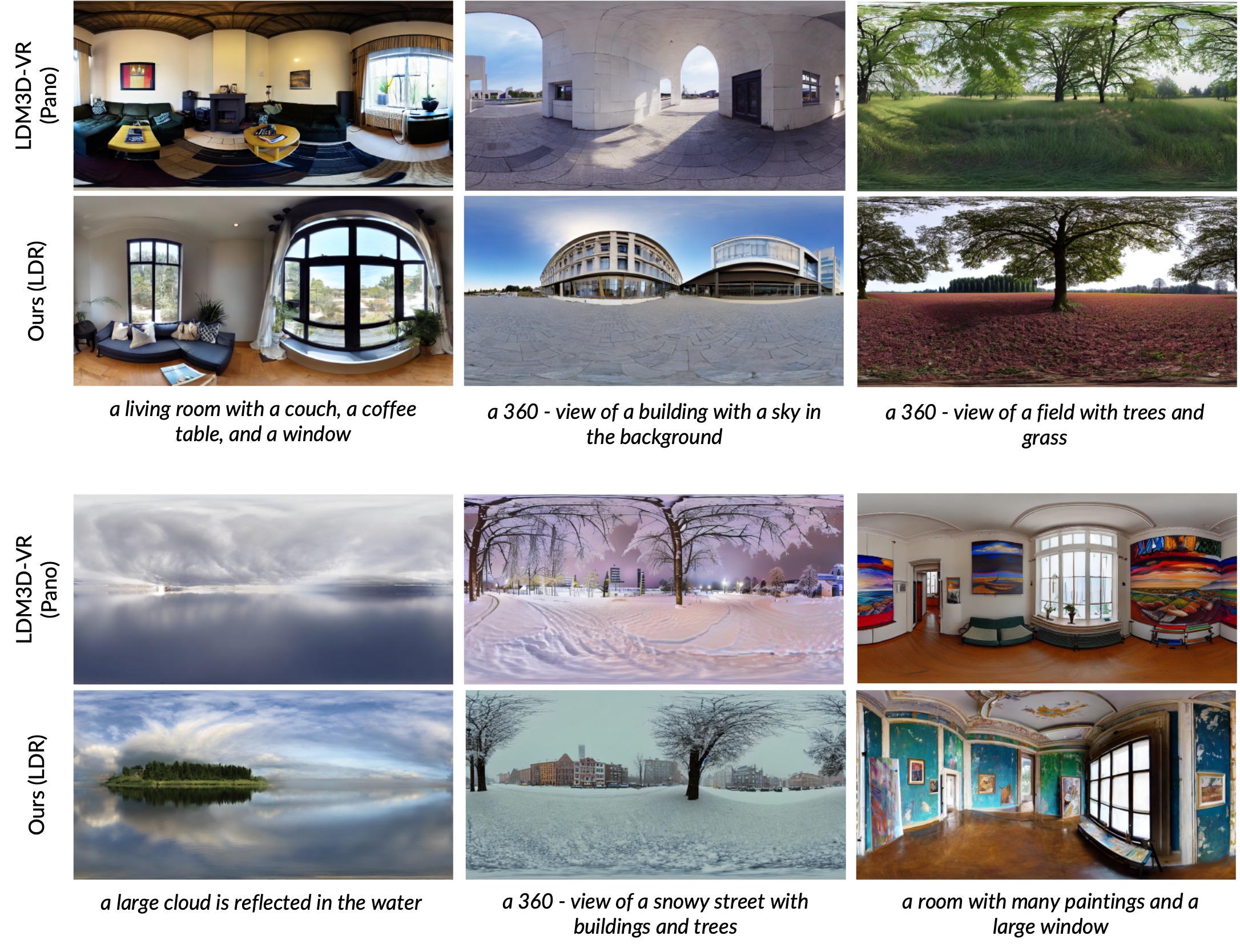}
    \caption{Our text to 360-degree LDR generative model synthesize realistic images for diverse prompts compared to the LDM3D-VR~\cite{stan2023ldm3d} method.}
    \label{fig:ldr-supp-01}
\end{figure*}

\begin{figure*}[!t]
    \centering
    \includegraphics[width=\textwidth]{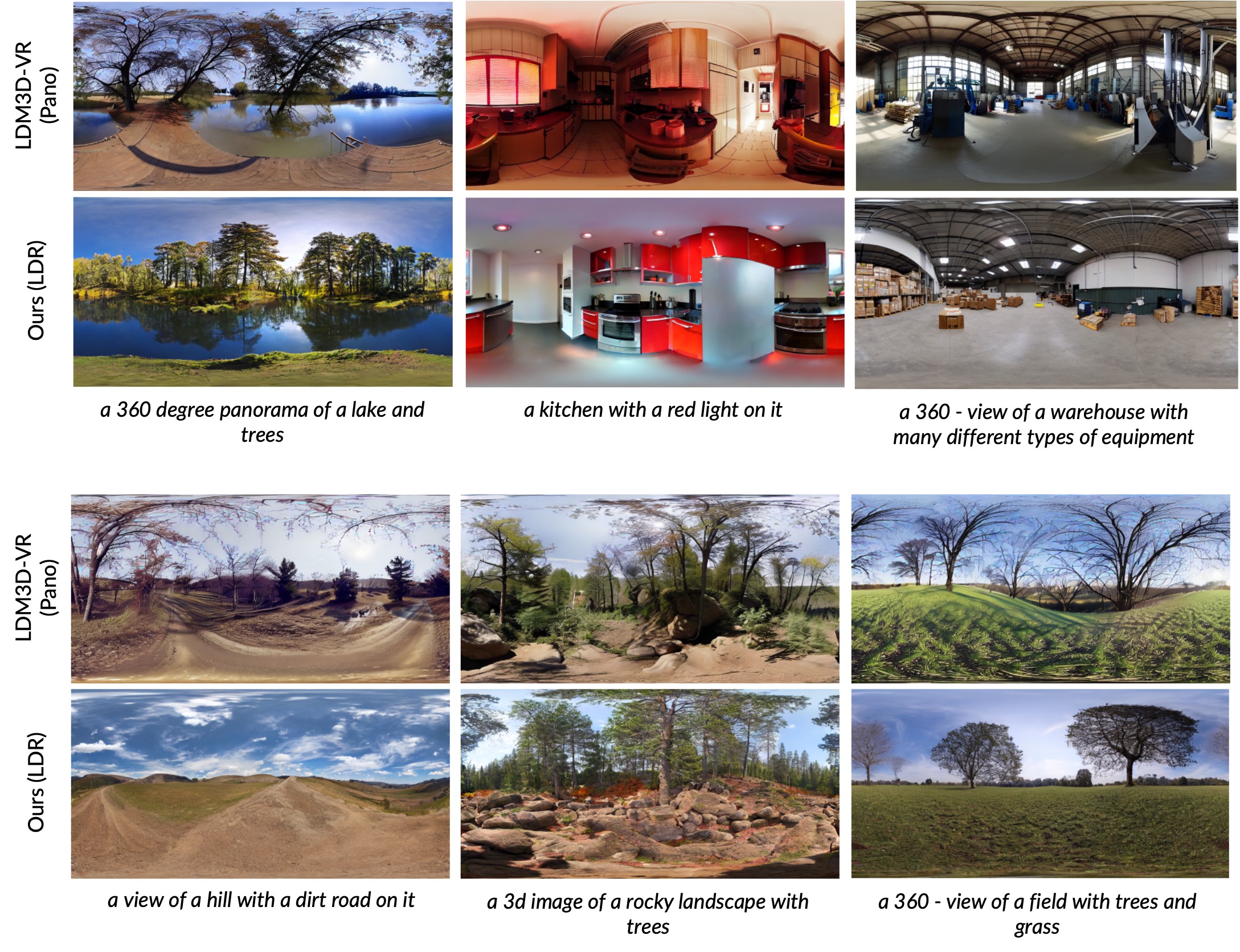}
    \caption{Ours text to 360-degree LDR generative model synthesis realistic images for diverse prompts compared to the LDM3D-VR~\cite{stan2023ldm3d} method.}
    \label{fig:ldr-supp-02}
\end{figure*}

\paragraph{LDR text to 360-degree generation} We compare additional qualitative realistic results from our LDR generative model with the LDM3D-VR approach for diverse text prompts in Fig.~\ref{fig:ldr-supp-01} and ~\ref{fig:ldr-supp-02}. Overall, our approach demostrates better prompt fidelity than the LDM3D-VR based on the generated images.


\clearpage

\end{document}